\documentclass[10pt,twocolumn,letterpaper]{article}

\usepackage{cvpr}
\usepackage{ulem}
\usepackage{times}
\usepackage{epsfig}
\usepackage{graphicx}
\usepackage{tabularx}
\usepackage{amsmath}
\usepackage{amssymb}
\usepackage{gensymb}
\usepackage{adjustbox}
\usepackage{array}
\usepackage{booktabs}
\usepackage{multirow}
\usepackage{rotating}
\usepackage{url}
\usepackage{caption}
\usepackage{subcaption}
\usepackage{color}

\newcolumntype{P}[1]{>{\centering\arraybackslash}p{#1}}

\graphicspath{ {Images/} }
\usepackage[pagebackref=true,breaklinks=true,letterpaper=true,colorlinks,bookmarks=false]{hyperref}

\cvprfinalcopy 

\def\cvprPaperID{0000} 

\ifcvprfinal\fi
\begin{document}

\title{Learning Dense Stereo Matching \\ for Digital Surface Models from Satellite Imagery}

\author{
Wayne Treible\textsuperscript{1,2} \quad  Scott Sorensen\textsuperscript{2} \quad Andrew D. Gilliam\textsuperscript{2} \quad Chandra Kambhamettu\textsuperscript{1} 
\quad Joseph L. Mundy\textsuperscript{2} \\ \\
\textsuperscript{1}University of Delaware, Newark, DE \\ 
\textsuperscript{2}Vision Systems Inc., Providence, RI \\
\tt\small\textsuperscript{1}wtreible@udel.edu \\
\tt\small\textsuperscript{2}\{wayne.treible, jlm\}@visionsystemsinc.com \\
\\
}

\maketitle

\begin{abstract}

Digital Surface Model generation from satellite imagery is a difficult task that has been largely overlooked by the deep learning community. Stereo reconstruction techniques developed for terrestrial systems including self driving cars do not translate well to satellite imagery where image pairs vary considerably. In this work we present neural network tailored for Digital Surface Model generation, a ground truthing and training scheme which maximizes available hardware, and we present a comparison to existing methods. The resulting models are smooth, preserve boundaries, and enable further processing. This represents one of the first attempts at leveraging deep learning in this domain.

\end{abstract}


\section{Introduction}
A Digital Surface Model (DSM) is a 3D representation of the surface of a region of terrain. In this work we will focus on modelling the surface of Earth, however DSM's may be generated for other planets or astronomical bodies such as asteroids. Digital Surface Models model the height of man made structures and foliage, unlike Digital Terrain Models(DTM's) which represent the height of the bare geologic surface without these structures. DSMs are  widely used for resource exploitation, architectural and civil engineering, and a variety of Geospatial Information tasks \cite{DEMInGIS}. DSMs and DTMs also enable a range of current and developing research topics. These include applications in change detection\cite{8354141,doi:10.1142/S1793431114500031,7361980,6471211}, crop measurements\cite{ecrs-2-05163}, weather analysis\cite{doi:10.1029/2002JD003296}, and scene understanding such as mapping dense historical cities\cite{10.1007/978-3-030-01762-0_17}. 

DSMs and DTMs can be produced by utilizing stereo reconstruction techniques on sets of satellite image pairs of a scene. For stereo matching commercial products often use a patch-based matching cost in addition to variations of semi-global block matching (SGM). A recent study compared two commercial products, OrthoEngine and RPC Stereo Processor (RSP)\cite{doi:10.1080/15481603.2018.1494408}. This work found RSP's modified hierarchical SGM method to be superior to OrthoEngine's hierarchical subpixel mean normalized cross correlation method. For a comparison of multi-view reconstruction using satellite images we refer the reader to \cite{7301292}

Satellites typically do not take wide baseline synchronized stereo imagery due to a number of physical and practical constraints. As a result, image pairs are captured by different satellites (or multiple passes of the same satellite) at different times. Additionally if arbitrary views can be used more images are available. This drastically complicates matching image points because illumination, environmental, and even large scale scene changes can occur. Differences in time of day and season change solar illumination angle and intensity. Cloud cover, snow, fog and recent rainfall can all affect image quality. Vegetation also undergoes seasonal change, and over the course of multiple years structures may be constructed, painted, or destroyed, changing the overall scene geometry and disrupting matching. 

\begin{figure}
\begin{subfigure}{.245\textwidth}
  \centering
  \includegraphics[width=0.89\linewidth]{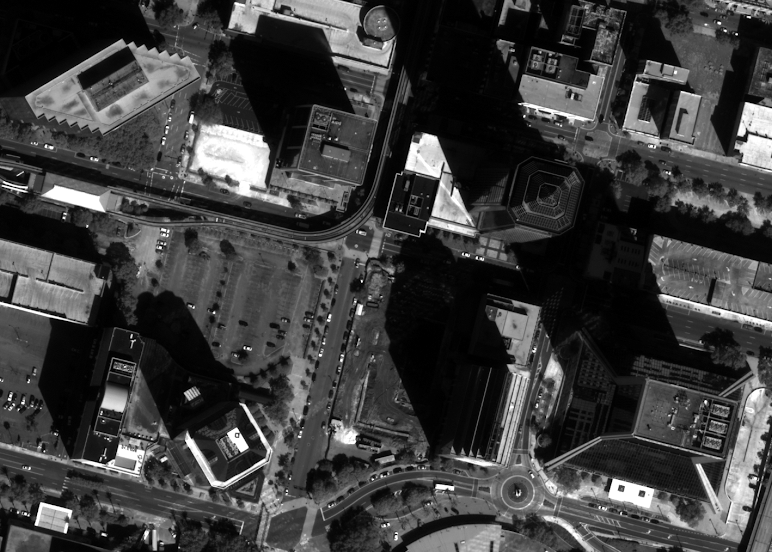}
  \caption{October 18 2014}
  \label{fig:sub1}
\end{subfigure}%
\begin{subfigure}{.245\textwidth}
  \centering
  \includegraphics[width=0.96\linewidth]{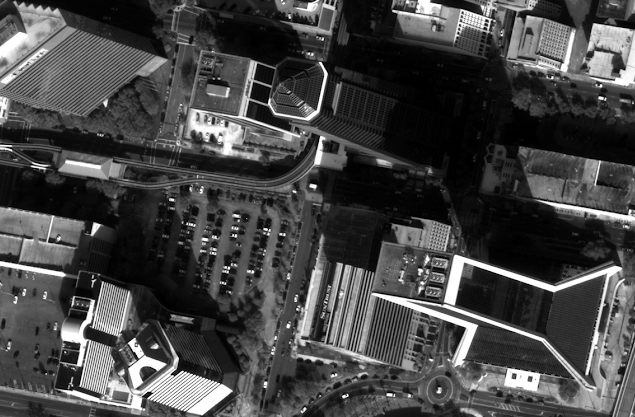}
  \caption{December 21 2015}
  \label{fig:sub2}
\end{subfigure}
\caption{Two satellite images of the same location showing a difference in viewing angle and scene lighting.}
\label{fig:test}
\end{figure}

While terrestrial stereo systems utilize a pinhole camera model and a defined world coordinate system, typically consisting of around 16 parameters. Satellite image projection is modeled using the Rational Polynomial Camera (RPC) model. RPC consist of 80 polynoimial and 10 scale and offset coefficients defining the projection of points at specific latitude, longitude, and elevation to image coordinates. Commercially available satellite images ship with RPC parameters, however better stereo reconstruction results can be achieved by further geocorrection \cite{Ozcanli2016}.


Recent works utilize neural networks to regress disparity from rectified stereo images. These methods formulate the problem of disparity calculation as a data-driven, end-to-end differential problem. Most of these methods are comprised of 4 main stages: Unary feature extraction, cost volume aggregation, regularization and smoothing, and finally the disparity regression itself. On the KITTI Stereo Benchmark, a street-view dataset to support autonomous driving, methods utilizing neural networks of this form occupy many of the top spots in all of the metrics. 

Traditionally, block-matching cost metrics such as SAD, census, and gradients are used to obtain correspondences for disparity calculation. For an in-depth summary of such methods, see \cite{Hassaballah}. Currently, the state-of-the-art for stereo image matching involves utilization of deep convolutional neural networks. End-to-end deep learning through training on large color image datasets was performed in Zagoruyko et al. \cite{Zagoruyko_2015_CVPR}. Dense stereo matching with deep neural networks was also performed in Luo et al. \cite{7780983}. At the time of writing, PSMNet was ranked second on the KITTI 2015 leaderboards by leveraging end-to-end learning and spatial pyramid pooling to increase the network's perceptive field \cite{chang2018pyramid}.

In this work, we utilize disparity-generating neural networks to solve the challenging problem of constructing DSMs from satellite imagery. Our contributions are as follows: 
\begin{enumerate}
\item Adapting Deep Neural Networks for disparity calculation on a new image domain (satellite imagery)
\item A method for generating ground truth disparity for satellite image pairs
\item A training scheme for combining satellite imagery and traditional stereo datasets
\end{enumerate}

\section{Acknowledgement}
Supported by the Intelligence Advanced Research Projects Activity (IARPA) via Department of Interior / Interior Business Center (DOI/IBC) contract number D17PC00288. The U.S. Government is authorized to reproduce and distribute reprints for Governmental purposes notwithstanding any copyright annotation thereon. The views and conclusions contained herein are those of the authors and should not be interpreted as necessarily representing the official policies or endorsements, either expressed or implied, of IARPA, DOI/IBC, or the U.S. Government. WorldView satellite images provided courtesy DigitalGlobe.

\begin{figure}[h]
  \includegraphics[width=0.5\textwidth]{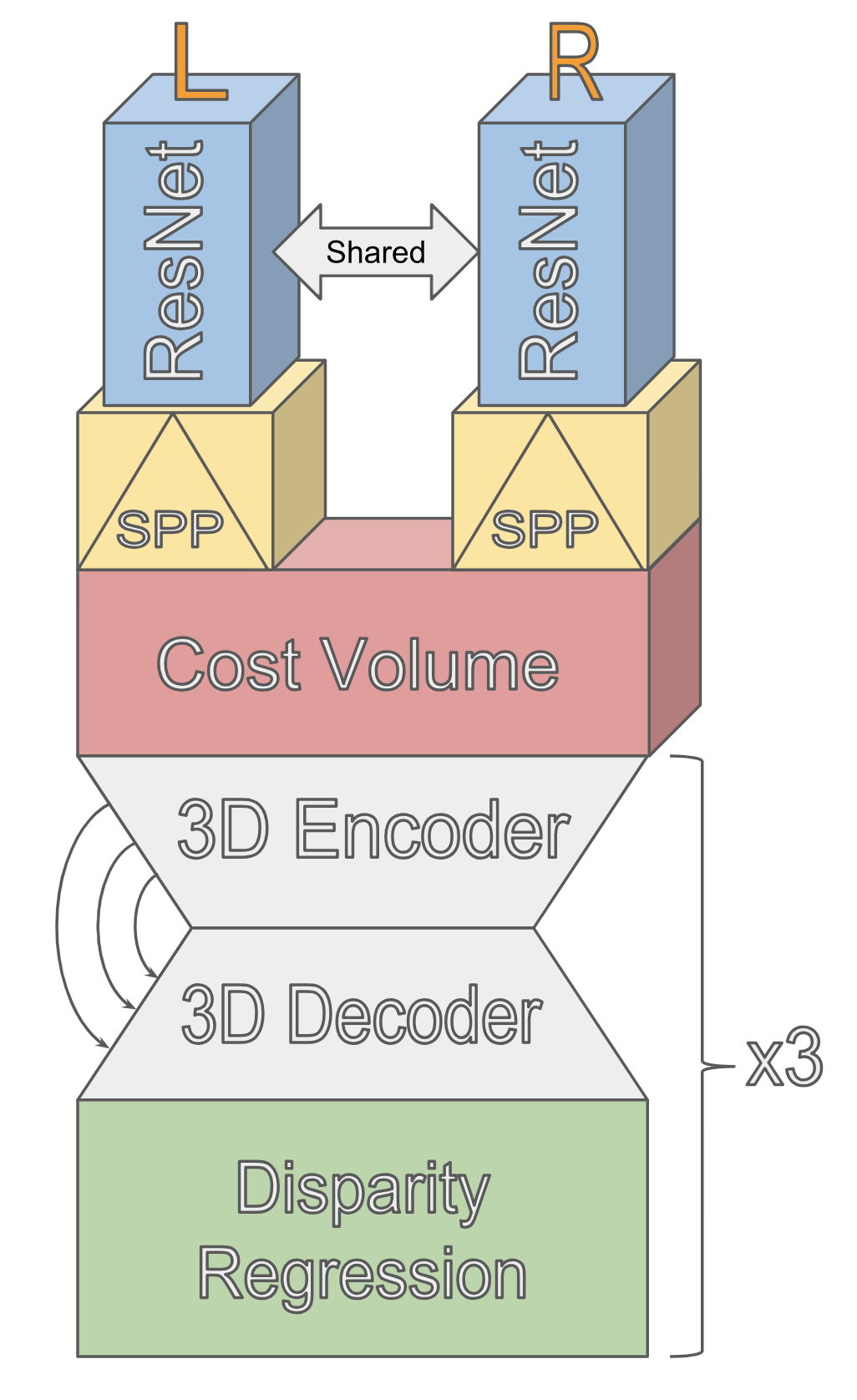}
  \centering
  \caption{A pictorial representation of the network architecture. In one configuration, ResNet-like branches share weights, and are each aggregated into a larger context with a Spatial Pyramid Pooling stage. }
\label{model_architecture}
\end{figure}

\renewcommand{\arraystretch}{1.5}
\begin{table}[]
\resizebox{\linewidth}{!}{%
\begin{tabular}{@{}c|c|c|c@{}}
\toprule
 \multicolumn{1}{c}{}  & \multicolumn{1}{c}{Layer Name} & \multicolumn{1}{c}{\begin{tabular}[c]{@{}c@{}}Layer Description\\ (conv shape, num filters, stride)\end{tabular}} & \begin{tabular}[c]{@{}c@{}}Layer Output Dimensions\\ (H x W x C)\end{tabular} \\ \toprule
1  & Input &  & H x W x 1C \\ 
2  & FeX\_InitialA & 3x3, 32, stride=2 & $\frac{1}{2}$H x $\frac{1}{2}$W x 32C \\
3  & FeX\_InitialB & 3x3, 32  & $\frac{1}{2}$H x $\frac{1}{2}$W x 32C \\
4  & FeX\_InitialC & 3x3, 32  & $\frac{1}{2}$ x $\frac{1}{2}$W x 32C \\
5  & FeX\_BlockStack0 & {[}3x3, 32{]} x 6 & $\frac{1}{2}$H x $\frac{1}{2}$W x 32C \\
6  & FeX\_BlockStack1 & {[}3x3, 64{]} x 32 & $\frac{1}{4}$H x $\frac{1}{4}$W x 64C \\
7  & FeX\_BlockStack2 & {[}3x3, 128{]} x 6 & $\frac{1}{4}$H x $\frac{1}{4}$W x 128C \\
8  & FeX\_BlockStack3 & {[}3x3, 128{]} x 6, dilation = 2 & $\frac{1}{4}$H x $\frac{1}{4}$W x 128C \\ \hline
9  & FeX\_SPP0 & \begin{tabular}[c]{@{}c@{}}64x64 average pooling\\ 3x3, 32\\ Linear/Cubic Upsampling \end{tabular} 
& $\frac{1}{4}$H x $\frac{1}{4}$W x 32C \\ 
&&&\\
10 & FeX\_SPP1 & \begin{tabular}[c]{@{}c@{}}32x32 average pooling\\ 3x3, 32\\ Linear/Cubic Upsampling\end{tabular}   & $\frac{1}{4}$H x $\frac{1}{4}$W x 32C \\
&&&\\
11 & FeX\_SPP2 & \begin{tabular}[c]{@{}c@{}}32x32 average pooling\\ 3x3, 32\\ Linear/Cubic Upsampling\end{tabular}   & $\frac{1}{4}$H x $\frac{1}{4}$W x 32C \\
&&&\\
12 & FeX\_SPP3 & \begin{tabular}[c]{@{}c@{}}32x32 average pooling\\ 3x3, 32\\ Linear/Cubic Upsampling\end{tabular}   & $\frac{1}{4}$H x $\frac{1}{4}$W x 32C \\
&&&\\
13 & FeX\_Concat & \begin{tabular}[c]{@{}c@{}}{[}FeX\_BlockStack1, \\ FeX\_BlockStack3,\\ FeX\_SPP0, FeX\_SPP1,
\\ FeX\_SPP2, FeX\_SPP3{]}\end{tabular} & $\frac{1}{4}$H x $\frac{1}{4}$W x 32C \\
&&&\\
14 & FeX\_LastConvA & 3x3, 128 & $\frac{1}{4}$H x $\frac{1}{4}$W x 32C \\
15 & FeX\_LastConvB & 1x1, 32 & $\frac{1}{4}$H x $\frac{1}{4}$W x 32C \\ \hline
16 & CostVolume & Concatenate disparity levels & $\frac{1}{4}$D x $\frac{1}{4}$H x $\frac{1}{4}$W x 32C \\ \hline
17 & PreHourglassBlock & {[}3x3, 32{]} x 4 & $\frac{1}{4}$D x $\frac{1}{4}$H x $\frac{1}{4}$W x 32C \\
18 & Hourglass0\_ConvA & 3x3, 64, stride=2 & $\frac{1}{8}$D x $\frac{1}{8}$H x $\frac{1}{8}$W x 64C \\
19 & Hourglass0\_ConvB & 3x3, 64 & $\frac{1}{8}$D x $\frac{1}{8}$H x $\frac{1}{8}$W x 32C \\
20 & Hourglass0\_ConvC & 3x3, 64, stride=2 & $\frac{1}{16}$D x $\frac{1}{16}$H x $\frac{1}{16}$W x 64C \\
21 & Hourglass0\_ConvD & 3x3, 64 & $\frac{1}{16}$D x $\frac{1}{16}$H x $\frac{1}{16}$W x 64C \\
22 & Hourglass0\_DeconvE & 3x3, 64, stride=2 & $\frac{1}{8}$D x $\frac{1}{8}$H x $\frac{1}{8}$W x 64C \\
23 & Hourglass0\_DeconvF & 3x3, 32, stride=2 & $\frac{1}{4}$D x $\frac{1}{4}$H x $\frac{1}{4}$W x 32C \\
24 & Hourglass1 & (See Above) & $\frac{1}{4}$D x $\frac{1}{4}$H x $\frac{1}{4}$W x 32C \\
30 & Hourglass2 & (See Above) & $\frac{1}{4}$D x $\frac{1}{4}$H x $\frac{1}{4}$W x 32C \\ \hline
36 & DispReg & \begin{tabular}[c]{@{}c@{}}Linear/Cubic Upsampling\\ Disparity Regression\end{tabular}                 & \begin{tabular}[c]{@{}c@{}}D x H x W\\ H x W\end{tabular} \\ 
\bottomrule
\end{tabular}
}
\caption{Network Description. This architecture builds upon PSMNet for the satellite image domain using dilated convolution at the end of the initial feature extraction stage, and 4 levels of SPP.}
\label{tab:network_description}
\end{table}

\section{Network Architecture}
\label{sec:architecture}

In this work, we present a deep neural network architecture for disparity regression from pairs of rectified images. This network has been adapted to the domain of satellite images. It builds upon Pyramid Stereo Matching (PSMNet) architecture \cite{chang2018pyramid}, which was a top performing method on the KITTI benchmark at the time of writing. A full description detailing the base architecture is given in Table \ref{tab:network_description}.

\subsection{Feature Extraction Modules (FeX)}

During our development we have experimented with multiple feature extraction (FeX) modules: A ResNet + Spatial Pyramid Pooling (SPP) Module, a ResNeXt + SPP Module, and a ResNet + Crosshair Pooling Module.

The first feature extraction module utilizes ResNet-like branches with shared weights coupled with a Spatial Pyramid Pooling (SPP) stage to aggregate large regions of context. The feature extraction begins with cascaded convolutional layers (3x3) that downsample the input image by a factor of two. This is followed by basic residual blocks (as in ResNet) and dilated convolution for extracting unary features across a very large receptive field. In satellite imagery, there is usually a very diverse set of man-made structures and foliage. This presents an interesting challenge on how to aggregate context information for feature maps. In the first module, feature context is combined with Spatial Pyramid Pooling. 

In SPP, average pooling over multiple scales captures a spatially-hierarchical context of the feature maps. Feature channels are compressed using 1x1 convolutions, then each pooled feature map is upsampled to the input feature map dimension with a user-defined choice of bilinear or cubic interpolation. The feature maps from each of the pyramid levels are then concatenated as the final output of this feature extraction module.

The second FeX module operates similarly to the first. In this module however, we utilize the improved residual blocks dubbed ResNeXt\cite{Xie2016}. Unary features are obtained from these residual blocks, and then aggregated using SPP in the same fashion as the first FeX module. 

The final FeX module uses the same unary feature extraction as the first FeX module. However, we introduce a new type of context pooling in order to capture different spatial context. We call this type of pyramid-style pooling "Crosshair" pooling, due to it's vertical and horizontal pooling bands. The motivation for this type of pooling is based on the epipolar geometry constraints of stereo matching, in which horizontal bands are pooled to capture vertical feature relationships perpendicular to the epipole, and vertical bands are pooled to capture horizontal relationships along the epipole. 

Of these feature extraction modules, the best qualitative performance was with the standard ResNet + SPP Module. For this reason, experimental results will be presented using this Feature Extraction module.
   
\subsection{Cost Volume Aggregation}
\label{costvol}

Each output of a FeX module is a 3-dimensional volume of feature maps that have encoded context information from Spatial Pyramid Pooling. These two volumes are aggregated with a correlation operation into a 4-dimensional cost volume. In this volume, each level represents a different disparity, with the right feature map is shifted and concatenated with the left or reference feature map. However, since the inputs have been downsampled, this volume is 1/4 the size of each spatial dimension and the newly added disparity dimension is 1/4 of the maximum disparity. The final 4D cost volume results in a shape of 1/4 Height x 1/4 Width x 1/4 Max Disparity x Number of Filters. 

\subsection{Smoothing and Regularization}
\label{hourglass}

The smoothing and regularization stage is performed by multiple stacked 3D hourglass CNNs. The encoder and decoder nature of an hourglass CNN captures additional spatial context information at varying scales. The use of 3D convolutions serve to not only aggregate feature information along the spatial dimensions, but also along the newly created disparity dimension of the cost volume. 

In our architecture, three 3D CNNs are stacked in a cascade, with each outputting a disparity map (via the disparity regression module) for intermediate supervision. This provides a type of iterative refinement to the smoothing and regularization aspects of this module. 

During training, all three hourglass outputs are fed to the Disparity Regression and are utilized in the loss calculation, with each being weighted by the empirically derived factors as in PSMNet ($loss_{total} = 0.5*loss_0 + 0.7*loss_1 + loss_2$). During testing and prediction, only the final disparity map is used, as it is the most refined output. 

\subsection{Disparity Regression}
\label{dispreg}

To regress disparity, we calculate the value of disparity of each individual pixel by taking the softmax of the cost along the disparity dimension at that specific pixel. We then sum each disparity value weighted by this softmaxed cost along the disparity axis of the 4D volume. This operation is typically called a soft-attention mechanism. This disparity regression is formalized as:

\[
  D(i,j) = \sum_{d=0}^{d_{max}} d \ast softmax(-C_{d}(i,j))
\]

That is, each pixel $(i,j)$ in the disparity map $D$ is the sum of the disparity $d$ multiplied by its softmaxed cost $C$ at the corresponding pixel across all disparity levels. There is strong evidence that utilizing this disparity regression is more robust than other methods for stereo networks \cite{KendallMDHKBB17}. 

Unlike in many other implementations, we do not end up completely squeezing the original 3D volume during the regression stage, and only totally regress a 2D map for loss calculations and prediction output. This allows us to utilize the original disparity costs in the cost volume for some downstream calculations and to better resolve ambiguities across multiple disparity maps that cover certain overlapping regions of interest.

\subsection{Network Modularity}
To facilitate matching in urban areas with tall buildings we have trained the network with a large disparity range (up to 352 maximum disparity). In practical terms this means splitting up the network into multiple pieces to enable training across multiple GPUs. To compute a 1 meter resolution DSM with the appropriate disparity range required approximately 20 GB of VRAM, and was trianed using two Nvidia Titan X GPU's on a workstation. Training at the full resolution of the input imagery (i.e. no initial downsampling convolution with half-resolution during cost volume aggregation) requires approximately 60 GB of VRAM.


\section{Dataset Overview}

The panchromatic grayscale images taken by the WorldView3 satellites have a higher metric resolution than the RGB imagery. There arises a small trade-off of 3-channel color information versus better spatial resolution. We elect to use the panchromatic grayscale imagery because high resolution DSMs are the end goal and previous works with SGM-based methods have proven grayscale imagery is sufficient. Therefore, all image data is first converted to grayscale so that 1-channel input image features can be learned. Our model is trained using a combination of existing terrestrial and synthetic stereo datasets, as well as a custom satellite dataset.

\subsection{Existing Stereo Datasets}

In many works on stereo matching, training is done with a combination of the four most popular stereo datasets: Sceneflow\cite{MIFDB16}, KITTI\cite{Geiger2012CVPR}, ETH3D\cite{schoeps2017cvpr}, and Middlebury\cite{ScharsteinHKKNWW14}.  Part of the training routine for our model is done using both Sceneflow and KITTI datasets.

The Sceneflow dataset is comprised of a collection of stereo pairs of synthetic scenes and contains dense, high-accuracy disparity maps. It also has the benefit of being one of the only datasets which contains a top-down view of a scene which is the most similar to the perspective of satellite imagery. The dataset contains over 35,000 images with disparity values of upwards of 500 pixels, which is useful for training the network to cover a large range of disparities. 

The KITTI dataset (2012 and 2015) is a popular stereo dataset which contains almost 400 images of street-view scenes with sparse ground-truth depth obtained from a LIDAR. The disparity maps in KITTI are therefore also sparse across the entirety of the image. KITTI data is useful because our satellite dataset also utilizes sparsely projected LIDAR points for ground truth, and presents similar difficulties.

\subsection{Satellite Stereo Dataset}
\label{satdata}

\begin{figure*}[ht]
\centering
\begin{subfigure}{.3\textwidth}
  \centering
  \includegraphics[width=0.98\linewidth]{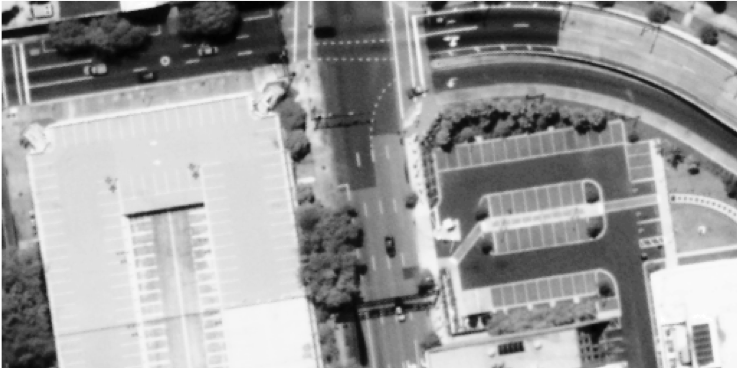}
\end{subfigure}
\begin{subfigure}{.3\textwidth}
  \centering
  \includegraphics[width=0.98\linewidth]{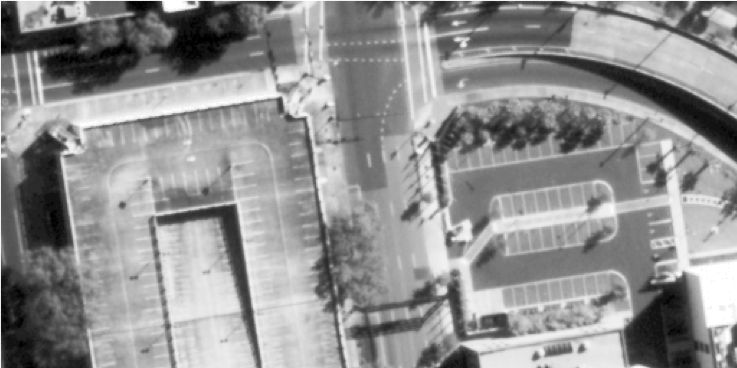}
\end{subfigure}
\begin{subfigure}{.3\textwidth}
  \centering
  \includegraphics[width=0.98\linewidth]{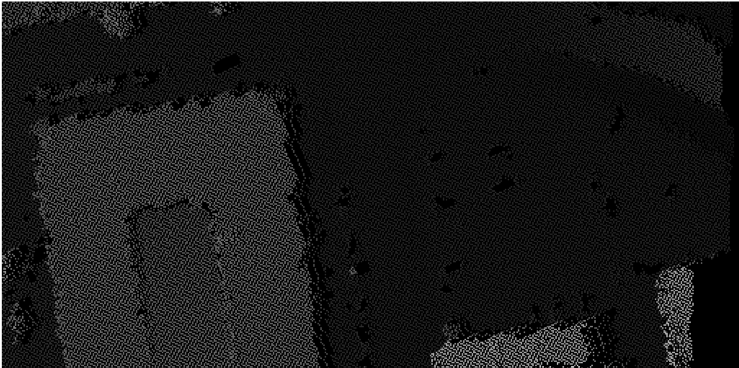}
\end{subfigure}
\caption{Sample stereo images and corresponding disparity map.}
\label{pair_and_disp}
\end{figure*}

Satellite imagery is a difficult domain for a variety of reasons. Images of the same scene often are captured at vastly different times, so image pairs often vary highly in lighting and shadow. Satellite images are also modeled using the RPC camera model and often need to be approximated by an affine camera to perform tasks such as rectification. Because of the unique domain of satellite imagery and associated cost, there aren't any sufficient training datasets of satellite image pairs with ground truth for training our network. Therefore, we have created our own satellite stereo dataset for training our models. 

We used a large set of DigitalGlobe WorldView 3 satellite imagery that has an approximate ground resolution of 31cm capturing a region of downtown Jacksonville, Florida. The satellite imagery spans a temporal range of approximately one year, with captures in multiple seasons and different times of day. For ground truth, an aircraft-mounted LiDAR was used to obtain high-accuracy depth of a subset of the imaged area at approximately 1 meter resolution and sampled at 50cm. 

Calculating ground truth disparity involved multiple camera model conversions, coordinate system conversions, transformations, and projections. Many of these operations leverage the suite of tools in VXL (Vision-Something-Libraries) which is a collection of C++ libraries and corresponding Python bindings for Computer Vision research and applications. A short summarization of the ground truth data generation is given below. 

To generate the rectified grayscale imagery, affine cameras are approximated for each of the RPC cameras. These approximated affine cameras are generated with an iterative optimization process and allow us to perform down-stream tasks such as rectification. The "scene," or region of interest, is cropped out from the larger satellite imagery. From there, random points on an approximated ground plane are projected into all combinations of pairs of cameras to find image correspondences. A homography is then calculated for each image in a pair, and the images are warped with bilinear interpolation to produce sets of rectified images. 

For ground truth, the LiDAR GeoTiff was converted from UTM (Universal Transverse Mercator) coordinates to WGS84 by using the GDAL library. WGS84 (World Geodetic System 1984) is a latitude, longitude, altitude format that dynamically changes over time with geologic events and represents a standard coordinate system for the Earth. These coordinates are then projected into each of the approximately 20 RPC  camera models (more specifically, the affine approximations of the camera models) that correspond to passes of the satellite over the scene. The generated pairs of homographies are then used to transform the LiDAR images to rectified image space. Each individual LiDAR point is used as the correspondence between the two rectified LiDAR images, and disparity is calculated by taking the difference along the epipole. 

Due to the sparseness of the LiDAR image compared to the satellite imagery, the ground truth disparity maps are also sparse and lack strong edges which presented some issues during training. The angle of the satellite images also caused a "bleed-through" effect of occluded ground points projecting between points on the roof of a building. To limit this impact, points with higher relative altitude were dilated, and points with an altitude less than the dilated mask (within a one meter tolerance) were removed. This further increased the jaggedness of the warped points, so bilinear interpolation was used to generate smoother, denser, disparity maps at the cost of slight error in the ground truth. Both the sparse disparity and the interpolated disparity were saved for different training procedures described in section \ref{training_procedures}.

We used these rectified pairs and associated ground truth to create images for training of size 256H x 512W by jointly sliding a window across all three images with a step size of 64 pixels in both the $x$ and $y$ directions. Image pairs that contained less than 40\% density in their ground truth images were not selected. The most common instances where this occurred were on the sub-images that contained a large portion of the side of a tall building (the top-down LiDAR ground truth contained no building sides) or along the edges of the ground truth region.

An additional challenge for determining training data was the uncertainty in relative satellite positioning. In traditional stereo camera setups, the relative position of the reference and target images is known (i.e. a "left camera" and a "right camera"). In order to make sure we have the canonical setup, only images with "positive" disparity -- a disparity value of $d$ represents the shift of a matching pixel in the form $reference(x,y) = target(x-d,y)$ -- were selected. Do to the nature of the direction and angle at which satellite images are captured, there were some instances with small "negative" disparity existed in the ground truth. If this amount was small (\textless10px), we simply shifted the target image by that amount and added that amount to all disparity values. If it was larger, we either flipped the images and checked again, or discarded that specific pair.

After all filtering, the final satellite stereo dataset used for training and testing contained over 30k pairs of stereo image pairs with corresponding ground truth disparity maps.

\begin{figure*}[ht]
\centering
\begin{subfigure}{.48\textwidth}
  \centering
  \includegraphics[width=0.89\linewidth]{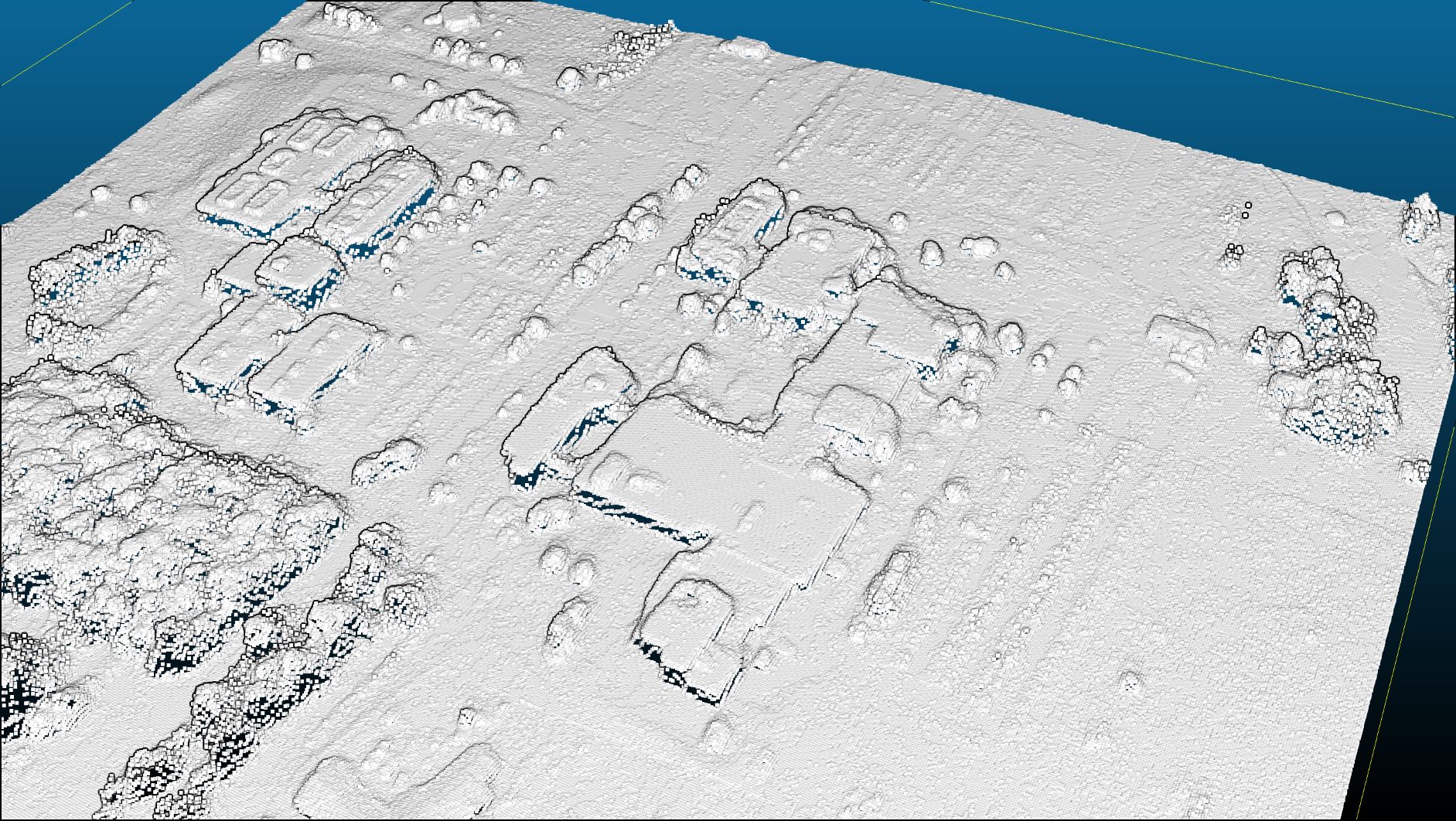}
  \caption{SGM Model}
  \label{fig:sub1_2}
\end{subfigure}%
\begin{subfigure}{.48\textwidth}
  \centering
  \includegraphics[width=0.89\linewidth]{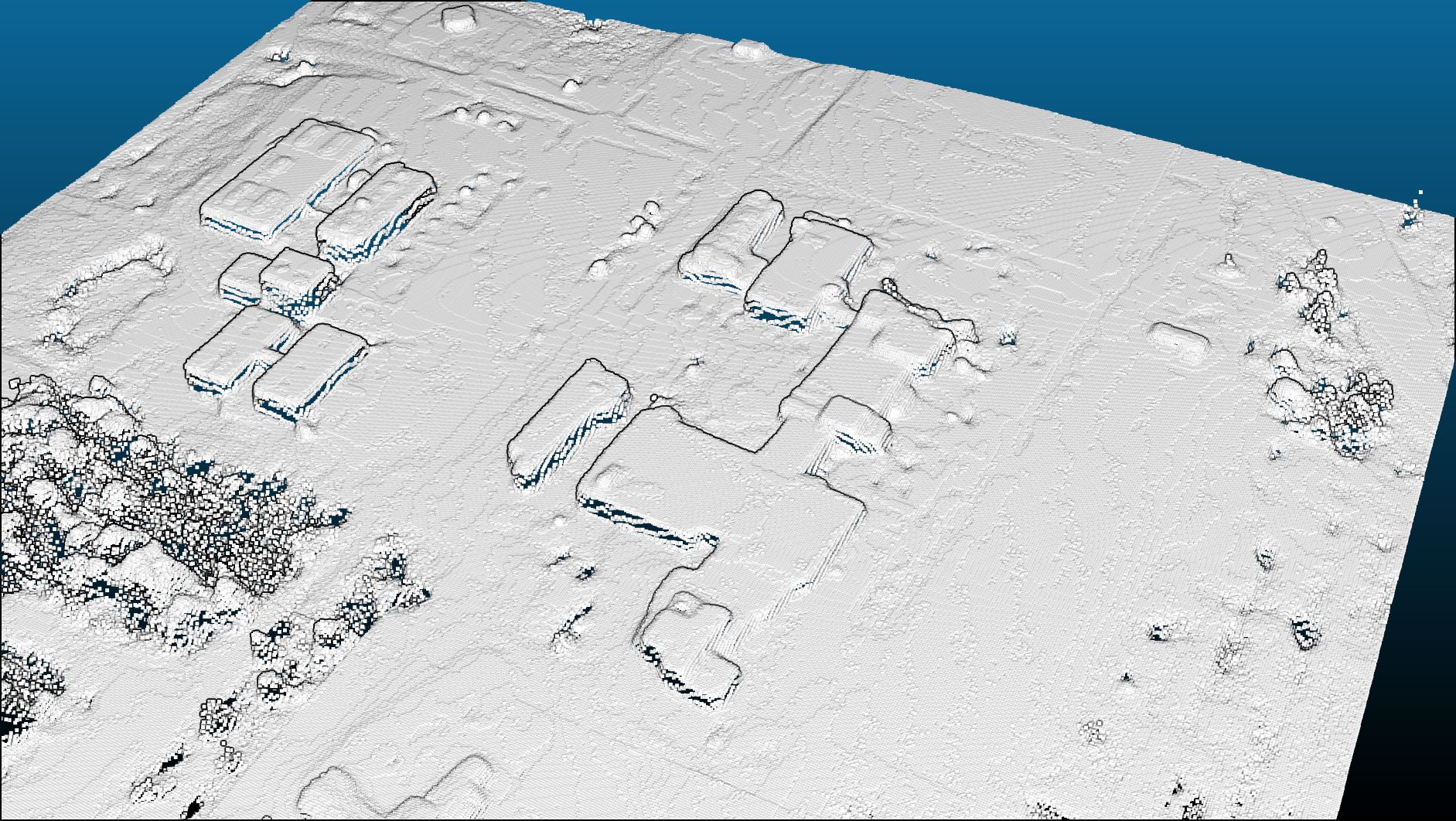}
  \caption{Neural Network Model}
  \label{fig:sub2_2}
\end{subfigure}
\caption{A shaded point cloud representation of the DSM created using the SGM process and Neural Network implementation.}
\label{fig:pointcloudfig}
\end{figure*}


\section{Experiments}

\subsection{Training Procedures}
\label{training_procedures}
Multiple training procedures were tested with a combination of different datasets which are briefly summarized below. All training procedures use normalized (zero mean \& unit variance) grayscale imagery with a vertical resolution of 256 pixels and a horizontal resolution of 512 pixels (256H x 512W x 1C). The Adam optimizer was used with the Keras parameter defaults of $\beta_1=0.9$ and $\beta_2=0.999$. The network was trained with a maximum disparity of 352 in order to account for the large disparity ranges caused by a mix of very tall buildings and smaller ground structures present in satellite imagery. All images were shuffled before training, and in each instance the networks were trained from scratch with the described procedures.

In two of the training procedures, we adopted the general training strategy used by most of the top networks. We first train the network on Sceneflow for 20 epochs at a learning rate of 0.001 followed by fine-tuning on a mix of KITTI and satellite imagery. 

In the first procedure, fine-tuning was performed with KITTI using a learning rate of 0.001 for 100 epochs followed by 10 epochs at a learning rate of 0.001 of training with satellite imagery. In this iteration, only a small set of the satellite imagery was used (approx. 15k image pairs). We decided to incorporate KITTI into our training procedure because it contained higher resolution ground truth than our satellite imagery. The set of satellite imagery contained multiple elevated roadways, a mix of small to moderately tall buildings, and some parks and vegetation. We call this training regimen "traditional transfer" as this approach is commonly used in adapting a network to a new domain.

In the second procedure we used a similar transfer learning style, instead removing KITTI imagery in favor of using more satellite images that contained denser urban regions with taller buildings. We tried this method because the disparity generated in the traditional transfer method appeared to be biased towards smaller disparity predictions when tall structures were present in the testing data. After the initial 20 epochs of Sceneflow training, 20 additional epochs with a learning rate of 0.001 were performed with both sets of satellite images shuffled together. We call this procedure "full transfer."

Due to the sparse nature of the ground truth of satellite images, many of the disparity maps that had been fine-tuned on satellite images would have very soft edges and rounded corners at the edges of buildings. We were also worried about over-fitting to urban scenes because our training data came from largely urban areas. In order to attempt to maintain the desired sharp edges in the final disparity map and increase network robustness, training was performed with a mix of the Sceneflow dataset (to learn hard-boundaries between disparity levels) and the 2 satellite image sets (to learn the domain-specific features and matching costs). In this procedure, 10k images were chosen from each set, for a total of 30k images. The images were shuffled together and trained at a constant learning rate of 0.001 for 20 epochs. We call this approach the "mixed" procedure.

\subsection{Evaluation of Training Procedures}
\label{sec:training_eval}

Evaluation of the training procedures was performed by calculating the accuracy on a small test set of satellite imagery that was compiled using a section of a scene containing the Wright Patterson Air Force Base near Dayton, OH. The evaluation set contained approximately 10,000 image pairs generated in the same fashion as previously described in \S \ref{satdata}. All three training procedures are evaluated using the same smooth L1 metric used in training and are presented in Table \ref{tab:training_eval}.

We determine that the "Full" procedure is the best generalizing procedure for new regions of satellite imagery. This isn't particularly surprising, as more unique satellite imagery in the training set should regularize the network better. 

\renewcommand{\arraystretch}{1.5}
\begin{table}[]
\resizebox{\linewidth}{!}{%
\begin{tabular}{l|llll}
Procedure   & \multicolumn{1}{l}{Regression Loss 1} & \multicolumn{1}{l}{Regression Loss 2} & \multicolumn{1}{l}{Regression Loss 3} & \multicolumn{1}{l}{Weighted Loss} \\ \hline
Traditional & 10.814                                 & 10.706                                 & 10.662                                 & 23.563                             \\ 
Full        & 10.139                                 & 9.940                                  & 9.916                                  & 21.943                             \\ 
Mixed       & 12.120                                 & 11.692                                 & 11.208                                 & 25.452                             \\ 
\bottomrule
\end{tabular}
}
\caption{Evaluation of Training Procedures. The smooth L1 metrics for each of the 3 disparity regressions as well as the final weighted value are reported on the test set. Here, we see that "Full" transfer is the best generalizing training procedure for a new region of satellite imagery.}
\label{tab:training_eval}
\end{table}

\begin{figure*}[ht]
\centering
\begin{subfigure}{.49\textwidth}
  \centering
  \includegraphics[width=0.95\linewidth]{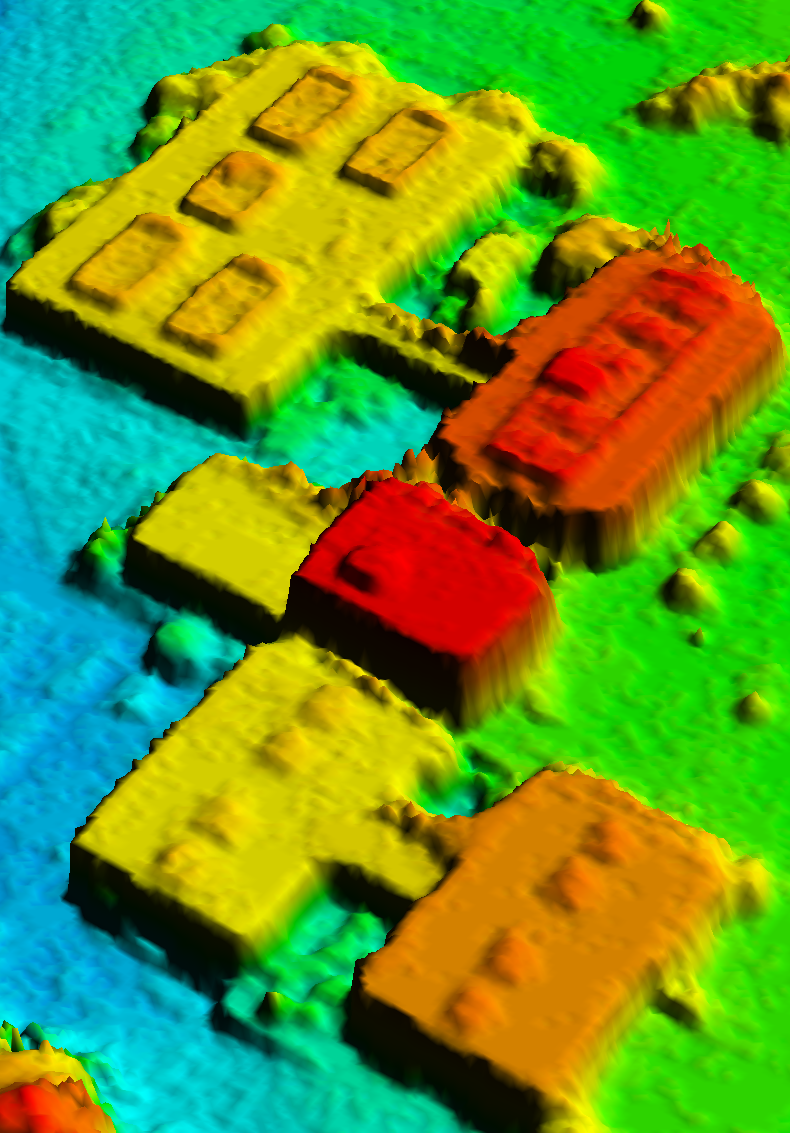}
\end{subfigure}
\begin{subfigure}{.49\textwidth}
  \centering
  \includegraphics[width=0.99\linewidth]{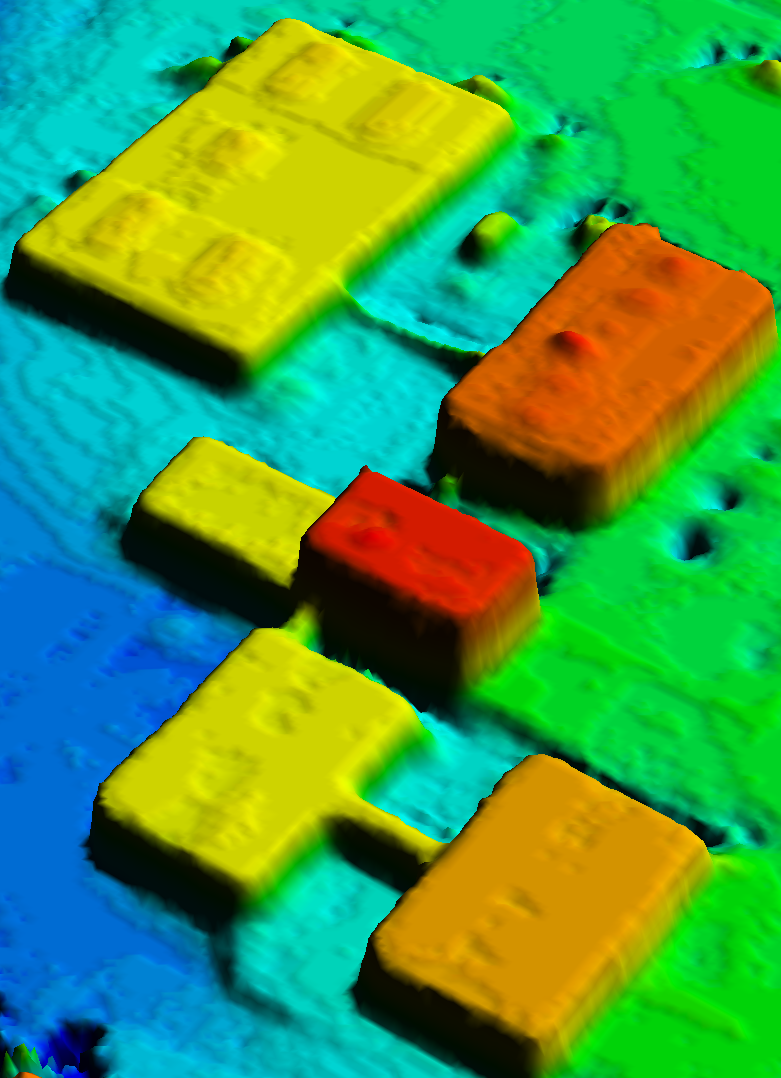}
\end{subfigure}
\caption{Reconstructed DSM using SGM (Left) and neural network implementation (Right) displayed using QT Modeler.}
\label{fig:qtfig}
\end{figure*}

\subsection{Digital Surface Model (DSM) Comparison}

DSMs were also constructed by merging pairwise disparity maps. Each disparity map is triangulated and reprojected to form a pairwise DSM. These pairwise DSM's are then combined using a fusion process that finds the most likely elevation for each point. This process was carried out for pairwise DSM's generated using both the network generated disparity as well as a SGM process. Results are presented in Figure \ref{fig:pointcloudfig} and Figure \ref{fig:qtfig}. 

Note the sharp edges in the Network generated DSM, which preserves straight edges on buildings. Figure \ref{fig:qtfig} highlights another advantage of this approach in how it performs in shaded regions. The SGM approach ofen leaves large amorphous regions particularly in the shadow side of the buildings (upper right in Figure \ref{fig:qtfig}). While building geometry and edges are preserved the network does suffer from a lack of detail in some places, for example the ventilation systems and piping on the roofs of these structures are over smoothed, potentially due to the downsampling in the network due to memory constraints.


\section{Conclusion}
\label{sec:conclusion}

In this work we have utilized a  neural network for disparity regression from satellite images. Matching pairs of satellite images is a difficult task because significant image variation. The architecture of this network has been developed for Digital Surface Model generation and trained using a scheme that enables dense matching and domain specific features. Resulting DSMs are smooth, have sharp edges, and preserve structural detail. 

The network was trained using a combination of existing and widely used stereo datasets and a custom dataset of approximately ~30k satellite image pairs. We have generated ground truth disparity maps for each image pair by projecting LiDAR point correspondences. Comparing the trained network to an existing robust implementation using Semi-Global Block Matching on an area never seen during the training of our network shows that the resulting DSM are detailed, smooth, and maintain sharp boundaries.


{\small
\bibliographystyle{ieee}
\bibliography{egbib}
}

\end{document}